\section{Post-conclusions discussion}
\label{sec:feedback}
\label{sec:responses}
\label{sec:appendix}

I received deeply thoughtful feedback on earlier versions of this article, including both reviewer feedback and comments elicited from people I thought would be likely to disagree with all or part of my argument. I note and respond to some of that feedback here, emphasizing that all representations of feedback and the responses to it are entirely my own and I am solely responsible for any errors.

\paragraph{\bf Is harmful LLM bias actually a thing? What's your evidence that existing mitigation methods aren't enough to prevent unmanageable user impact?} I was surprised to see this question asked, but then after digging into the literature and asking a number of well informed experts, I found it was surprisingly difficult to find any systematic, empirical studies on actual, real-world harms caused by LLMs or the data on which they are trained. A great number of studies demonstrate empirically that harmful LLM biases \emph{exist}, but this is done exclusively, as far as I can tell, via \emph{in vitro} methods --- as a typical example, \citet{hirota2022quantifying}, in a discussion of visual question answering datasets, write, ``Our findings suggest that \emph{there are dangers} associated to using VQA datasets without considering and dealing with the \emph{potentially harmful} stereotypes'' (my emphasis). Similarly, the results of \citet{hofmann2024dialect}, to which I have referred frequently in the main article, ``demonstrate that dialect prejudice \emph{has the potential for harmful consequences} by asking language models to make \emph{hypothetical} decisions about people'' (my emphasis). So this is not just a question for the main article; it's a question for the entire, very active community of people funding and conducting research on bias prevention and mitigation.

I do believe that it is a fair question to ask, from a scientific perspective. At the same time, I would note that in most other domains where new developments have a large impact, evidence of \emph{potential} harm is sufficient motivation for a far higher degree of caution than we are seeing with LLMs. For example, if a promising drug shows evidence of potential harm in \emph{in vitro} studies or in clinical trials, this can often lead to delay, reformulation, or outright abandonment of the drug before advancing to broader studies or widespread deployment \citep[e.g.][]{bass2004origins}. You won't even find a toaster oven in a Walmart in the U.S. without an Underwriter's Laboratory (UL) safety testing label. And yet technology companies are marketing products that dupe unwitting lawyers into submitting fake legal briefs to judges \citep{schwartz2023}, generate ``thinspiration'' images for anorexics and bulimics \citep{fowler2023ai} and advise them on ways to lose weight \citep{neda2023}, that sexually harrass users \citep{cole2023my}, and that lead fragile people to take their own lives \citep{graber2023world,roose2024}.\footnote{Sewell Setzer~III, a young teenager with no history of suicide risk factors, spent months talking with a ``Daenerys Targaryen'' chatbot from Character.ai, pulled away from his real-world connections, and became convinced that her ``world'' was actually the real world and he wanted to be in it.  In conversations where Sewell expressed suicidal ideation, the character would give responses like ``don't you dare talk like that'', but ``she'' never broke character or warned anyone outside the conversation. Readers can assess the last conversation with ``Daenerys'', in February 2024, for themselves. \emph{Sewell: “I miss you”. Daenerys: “I miss you too''. Sewell:“I’ll come home to you. I love you so much, Danae.” Daenerys: “I love you, too. Please come home to me as soon as possible, my love.” Sewell: “What if I could come home right now?” Daenerys: “Please do, my sweet king.”}  Moments later Sewell killed himself using his stepfather's handgun \citep{hardfork2024}.
  
  Noam Shazeer and Daniel de Freitas, the founders of Character.ai, previously worked at  Google but left to create Character.ai because ``Google was this bureaucratic company that had all these strict policies, and it was very hard to launch anything, quote, 'fun''' \citep{hardfork2024}.

  Google re-hired Shazeer and de Freitas in August 2024 and will be licensing Character.ai's LLM technology. The company has been reported to be worth \$2.5 billion \citep{metz2024}.}
While these may not be examples of bias, strictly speaking, the underlying issues are closely related: first, the troubling examples above arose via the same process I have discussed, in which a system's dominant probabilistic structure minimized prediction error for human-generated text; and second,  any ``guard rail'' efforts to rein in normatively unacceptable or even dangerous system behaviors in the pretrained models must trade off safety against the overriding goal of making sure that systems continue to ``work well'' as measured by benchmarks, user adoption, and corporate valuations.

\paragraph{\bf How do we actually know that the relevant distinctions are not discovered distributionally by LLMs?}  It is true that the present article does not provide a \emph{proof} that pre-trained LLMs are not somehow encoding the necessary distinctions I have highlighted. This can be compared to \citet{bouyamourn-2023-llms}, for example, who offer a formal proof that LLMs operating under plausibly standard assumptions must hallucinate. However, I have presented a careful argument laying out why LLMs could not distinguish among facts of conventional meaning versus contingent/normative and contingent/non-normative propositions, and any claim to the contrary needs to be supported either by a convincing counter-argument or by valid evidence --- evidence that, per my discussion, cannot merely involve asking the model to make specific distinctions behaviorally, since underlying biases can remain despite their absence in overt  direct-response behavior \citep{hofmann2024dialect}. If this article's argument is plausible enough to give rise to community level  back-and-forth discussion of the arguments and counter-arguments, then I will have succeeded in my mission. I also conjecture that the proof of \citet{bouyamourn-2023-llms} may be a promising avenue for actual formal results about harmful bias, given that factuality and conventional meaning are closely related.

\paragraph{\bf It was already obvious that harmful biases are baked into LLMs} Interestingly, a view opposite to the one above is also argued: that I am just stating the obvious, because there is no such thing as an objective ``view from nowhere'' \citep{nagel1989view} --- e.g. see \citet{sep-scientific-objectivity,kaeser2020positionality}. On this reading, subjective bias is inevitable for both people and machines, and RLHF and similar mitigations are merely strategies for substituting one set of biases for another. I am sympathetic to this view, and strongly endorse some of its key implications \citep[e.g. see][]{blodgett-etal-2020-language}, particularly that technological work related to bias needs to connect much more directly with relevant expertise in other disciplines, and that researchers and developers need to formulate and communicate explicit normative reasoning, rather than relying on small, highly influential groups of technologists basing their widely-deployed technologies on \emph{ad hoc} characterizations of harm \citep[e.g.][]{ouyang2022training} or on informal attempts to ``gather a thoughtful set of principles'' \citep[e.g.][]{claude}.

\paragraph{\bf What about people? People are biased, too}  This response has intuitive appeal, but on closer inspection the analogy is shallow and doesn't hold. First, contrary to most of human history, today (at least when we are at our best) we view harmful biases as societally important. So to turn the question around: as long as someone is really productive and useful, does that mean we needn't prioritize countering their racism, misogyny, antisemitism, etc. as highly as we prioritize their productivity?  Unlike human bias, LLM bias does not \emph{have} to be Society-complete. It’s fact about their design, and designs can be changed.

Second, with human biases we have a much clearer idea what to expect --- for better or worse --- and therefore we have experience countering it. As a striking example, \citet{anderson2022talk} discuss ``the Talk'' that many Black parents in the U.S. have with children grounded in their knowledge about what to expect in interactions with the police, emphasizing the challenge of ``alerting their children of possible harm while also not villainizing every member of law enforcement their child may encounter''.\footnote{For some fascinating research on how non-obvious biases can affect human decision-making in law enforcement, see \citet{fridman2019applying}. They draw on research in neuroscience to demonstrate how catastrophic outcomes can arise as a result of poor predictions that arise from generalizations in people's internal models.}  A wealth of research, experience, and mitigation strategy has evolved over decades informed by the understanding of sources of human error \citep{reason1990human}, including cognitive and other biases in high-stakes domains like medicine \citep{croskerry2002achieving,ng2025clinical}. In contrast, again as brought out by \citet{hofmann2024dialect}, \citet{betley2025emergent}, and other work, we have little idea what expect of LLMs in terms of what's going on under the hood, little user experience with recognizing it and dealing with it, and no well established, reliable ways to find out.

\paragraph{\bf What empirical evidence \emph{would} convince you that LLMs are making relevant distinctions/not encoding harmful biases in their representations?} My view is that this shouldn't be solely an empirical question---at least not in the current benchmark-style mode of empirical evaluation, where improving scores for black box systems generally takes priority over improving scientific understanding. In most other settings, our confidence in a solution's safety rests partly on empirical testing and crucially also on our \emph{understanding of} the thing we are testing. For example, if a new immunotherapy is being introduced for cancer, using the body's own immune system to combat the disease, our knowledge about the treatment's underlying mechanisms tells us that we must look closely at autoimmune-related risks \citep{dhodapkar2019autoimmune}. 

If my primary argument is valid, LLMs' representational spaces include generalizations derived from innumerable distributional $n^{th}$-order relationships, both word-to-word and among representations themselves. In the absence of a reasonable understanding of underlying mechanisms, we've seen that the effects of those generalizations have a way of stubbornly persisting even for problematic biases we know to look for \cite{hofmann2024dialect,serrano-etal-2023-stubborn}. I therefore think it likely that the current mode of empirical testing, followed by attempting to mitigate remaining problems, followed by further empirical testing, etc.\ will amount to a never-ending game of whack-a-mole \citep[e.g.\ see][]{roth2024google}, something strongly reinforced by the recent surprising results of \citet{betley2025emergent}. On the other hand, I conjecture that if pure distributionalism were leavened with a degree of interpretable structure, e.g.\ distinguishing conventional from distributional aspects of representation (as floated in Section~\ref{sec:howtofix}), it might be possible to operationalize principles of the form, ``inferences related to property or goal A must[n't] rely on representational properties in category B'', where human-stated categories and principles arise from transparent, well informed normative reasoning \citep{blodgett-etal-2020-language}. Increased transparency of representations and mechanisms would then permit more informative empirical tests.

\paragraph{\bf Mightn't that lead to less useful models?} Quite possibly, at least for some period of time. But the terms of the LLM ``arms race'' launched in November 2022 \citep{grossman2024tech}, prioritizing utility, market share, and rapid technological evolution, are a choice, not a necessity. One can imagine an alternate history in which considerations of factuality, harmful bias, democratization of development, and more \citep{Bommasani2021FoundationModels} had originally played a role on par with those other priorities in discussions among the decision-makers developing industry-scale LLMs, as their potential became clear. That apparently didn't happen, but even now the role of those considerations going forward is still a choice, not a necessity. Unfortunately it is a choice over which very few of us have any influence, but my hope is that this article might help move the needle at least a little bit.

\paragraph{\bf C'mon man, don't be such a downer}  \emph{``The conclusion seems to suggest there is nothing we can do about bias ... If that's your 'final answer,' can you spin it more positively with something more like: 'Don't worry; be happy'''. ``If I were an LLM developer, neck deep in engineering challenges, I'm not sure I would have time for philosophers either''. ``'A lot of what’s in people’s heads sucks' ...  implies an unjustifiably pessimistic view of human thought as a whole''. ``We [the *ACL audience] no longer have the patience for a long discussion''.} --Anonymous Reviewers

I've argued that that we as a research community, and the dominant industry players with their planned 2025 spend of \$275 billion \citep{rattner2025ai}, are investing in a technological approach where current approaches to harmful bias cannot succeed. In principle. That's a tough pill to swallow. However, we can't address issues we don't discuss, so we need to work hard to make room for careful, thoughtful discussion, especially given that in the U.S., prioritizing speed of development over product safety has now graduated from corporate practice to national policy \citep{vance2025ai}. Against that backdrop, writing this article is an act of profound optimism.

\paragraph{\bf You are assuming \underline{\ \ \ \ \ } about LLMs, and \underline{\ \ \ \ \ } might not always be true} This is a fair point. For example, one of the assumptions implicit in my discussion is that LLMs are ``general purpose'', in the sense of a single underlying pre-training corpus yielding a single pre-trained model across use cases and users.  Another is that pre-training involves a huge, essentially non-systematic sample of human text, as opposed to selected materials that may give rise to less bias.  Yet another is that pre-training lacks non-textual evidence to support grounding \citep{harnad1990symbol,bender2020climbing,julianmichael}.  I believe my assumptions are consistent with the way LLMs are most widely used today.  To the extent that these things evolve, my conclusions about the inevitability of bias could also evolve.

One reviewer kindly encourages an even more confident response, noting: ``Under the assumptions the paper makes, its argument is valid; if someone develops an LLM that violates these assumptions, the paper makes no claims about such an LLM.''  This helpful comment emphasizes my most important point, which is that, if we want to tackle the problem of bias, what we should be focused on first is not mitigation, but rather interrogation of LLMs' foundations. Again, if this article contributes to broader, more thoughtful discussion \emph{of} the assumptions underlying LLMs and their relationship to harmful bias --- rather than everyone by default adopting whatever assumptions the dominant language models carry with them --- I will view it as a success.

\ignore{

\starttwocolumn
\bibliography{resnik_bias.bib}

\begin{thebibliography}{92}
\expandafter\ifx\csname natexlab\endcsname\relax\def\natexlab#1{#1}\fi

\bibitem[{Adcock and Collier(2001)}]{adcock2001measurement}
Adcock, Robert and David Collier. 2001.
\newblock Measurement validity: A shared standard for qualitative and
  quantitative research.
\newblock \emph{American political science review}, 95(3):529--546.

\bibitem[{Anderson(2008)}]{anderson2008end}
Anderson, Chris. 2008.
\newblock The end of theory: The data deluge makes the scientific method
  obsolete.
\newblock \emph{Wired magazine}, 16(7):16--07.

\bibitem[{Anderson, O’Brien~Caughy, and Owen(2022)}]{anderson2022talk}
Anderson, Leslie~A, Margaret O’Brien~Caughy, and Margaret~T Owen. 2022.
\newblock “{T}he {T}alk” and parenting while {B}lack in {A}merica:
  Centering race, resistance, and refuge.
\newblock \emph{Journal of Black Psychology}, 48(3-4):475--506.

\bibitem[{Anthropic(2023)}]{claude}
Anthropic. 2023.
\newblock {C}laude’s constitution.
\newblock \url{https://www.anthropic.com/news/claudes-constitution}. [Online.
  Accessed May 14, 2024].

\bibitem[{Bass, Kinter, and Williams(2004)}]{bass2004origins}
Bass, Alan, Lewis Kinter, and Patricia Williams. 2004.
\newblock Origins, practices and future of safety pharmacology.
\newblock \emph{Journal of Pharmacological and Toxicological Methods},
  49(3):145--151.

\bibitem[{Bender and Koller(2020)}]{bender2020climbing}
Bender, Emily~M and Alexander Koller. 2020.
\newblock Climbing towards {NLU}: On meaning, form, and understanding in the
  age of data.
\newblock In \emph{Proceedings of the 58th annual meeting of the Association
  for Computational Linguistics}, pages 5185--5198.

\bibitem[{Bengio, Courville, and Vincent(2013)}]{bengio2013representation}
Bengio, Yoshua, Aaron Courville, and Pascal Vincent. 2013.
\newblock Representation learning: A review and new perspectives.
\newblock \emph{{IEEE} Transactions on Pattern Analysis and Machine
  Intelligence}, 35(8):1798--1828.

\bibitem[{Betley et~al.(2025)Betley, Tan, Warncke, Sztyber-Betley, Bao, Soto,
  Labenz, and Evans}]{betley2025emergent}
Betley, Jan, Daniel Tan, Niels Warncke, Anna Sztyber-Betley, Xuchan Bao,
  Mart{\'\i}n Soto, Nathan Labenz, and Owain Evans. 2025.
\newblock Emergent misalignment: Narrow finetuning can produce broadly
  misaligned {LLM}s.
\newblock \emph{arXiv preprint arXiv:2502.17424}.

\bibitem[{Blei, Ng, and Jordan(2003)}]{blei2003latent}
Blei, David~M, Andrew~Y Ng, and Michael~I Jordan. 2003.
\newblock Latent {Dirichlet} allocation.
\newblock \emph{Journal of Machine Learning Research}, 3(Jan):993--1022.

\bibitem[{Blodgett et~al.(2020)Blodgett, Barocas, Daum{\'e}~III, and
  Wallach}]{blodgett-etal-2020-language}
Blodgett, Su~Lin, Solon Barocas, Hal Daum{\'e}~III, and Hanna Wallach. 2020.
\newblock Language (technology) is power: A critical survey of {``}bias{''} in
  {NLP}.
\newblock In \emph{Proceedings of the 58th Annual Meeting of the Association
  for Computational Linguistics}, pages 5454--5476, Association for
  Computational Linguistics, Online.

\bibitem[{Bommasani et~al.(2021)Bommasani, Hudson, Adeli, Altman, Arora, von
  Arx, Bernstein, Bohg, Bosselut, Brunskill
  et~al.}]{Bommasani2021FoundationModels}
Bommasani, Rishi, Drew~A Hudson, Ehsan Adeli, Russ Altman, Simran Arora, Sydney
  von Arx, Michael~S Bernstein, Jeannette Bohg, Antoine Bosselut, Emma
  Brunskill, et~al. 2021.
\newblock On the opportunities and risks of foundation models.
\newblock \emph{ArXiv}.
\newblock Preprint arXiv:2108.07258.

\bibitem[{Borg and Fisher(2021)}]{borg2021semantic}
Borg, Emma and Sarah~A. Fisher. 2021.
\newblock Semantic content and utterance context: a spectrum of approaches.
\newblock In Piotr Stalmaszczyk, editor, \emph{The {C}ambridge Handbook of the
  Philosophy of Language}, Cambridge Handbooks in Language and Linguistics.
  Cambridge University Press.

\bibitem[{Bouyamourn(2023)}]{bouyamourn-2023-llms}
Bouyamourn, Adam. 2023.
\newblock Why {LLM}s hallucinate, and how to get (evidential) closure:
  Perceptual, intensional, and extensional learning for faithful natural
  language generation.
\newblock In \emph{Proceedings of the 2023 Conference on Empirical Methods in
  Natural Language Processing}, pages 3181--3193, Association for Computational
  Linguistics, Singapore.

\bibitem[{Caliskan, Bryson, and Narayanan(2017)}]{caliskan2017semantics}
Caliskan, Aylin, Joanna~J Bryson, and Arvind Narayanan. 2017.
\newblock Semantics derived automatically from language corpora contain
  human-like biases.
\newblock \emph{Science}, 356(6334):183--186.

\bibitem[{Card et~al.(2022)Card, Chang, Becker, Mendelsohn, Voigt, Boustan,
  Abramitzky, and Jurafsky}]{card2022computational}
Card, Dallas, Serina Chang, Chris Becker, Julia Mendelsohn, Rob Voigt, Leah
  Boustan, Ran Abramitzky, and Dan Jurafsky. 2022.
\newblock Computational analysis of 140 years of us political speeches reveals
  more positive but increasingly polarized framing of immigration.
\newblock \emph{Proceedings of the National Academy of Sciences},
  119(31):e2120510119.

\bibitem[{Casper et~al.(2023)Casper, Davies, Shi, Gilbert, Scheurer, Rando,
  Freedman, Korbak, Lindner, Freire, Wang, Marks, Segerie, Carroll, Peng,
  Christoffersen, Damani, Slocum, Anwar, Siththaranjan, Nadeau, Michaud, Pfau,
  Krasheninnikov, Chen, Langosco, Hase, Bıyık, Dragan, Krueger, Sadigh, and
  Hadfield-Menell}]{casper2023open}
Casper, Stephen, Xander Davies, Claudia Shi, Thomas~Krendl Gilbert, Jérémy
  Scheurer, Javier Rando, Rachel Freedman, Tomasz Korbak, David Lindner, Pedro
  Freire, Tony Wang, Samuel Marks, Charbel-Raphaël Segerie, Micah Carroll,
  Andi Peng, Phillip Christoffersen, Mehul Damani, Stewart Slocum, Usman Anwar,
  Anand Siththaranjan, Max Nadeau, Eric~J. Michaud, Jacob Pfau, Dmitrii
  Krasheninnikov, Xin Chen, Lauro Langosco, Peter Hase, Erdem Bıyık, Anca
  Dragan, David Krueger, Dorsa Sadigh, and Dylan Hadfield-Menell. 2023.
\newblock Open problems and fundamental limitations of reinforcement learning
  from human feedback.
\newblock \emph{Computing Research Repository}, arXiv:2307.15217.

\bibitem[{Church and Hanks(1990)}]{church1990word}
Church, Kenneth and Patrick Hanks. 1990.
\newblock Word association norms, mutual information, and lexicography.
\newblock \emph{Computational linguistics}, 16(1):22--29.

\bibitem[{Church and Mercer(1993)}]{church1993introduction}
Church, Kenneth and Robert~L Mercer. 1993.
\newblock Introduction to the special issue on computational linguistics using
  large corpora.
\newblock \emph{Computational linguistics}, 19(1):1--24.

\bibitem[{Clark(2015)}]{clark2015surfing}
Clark, Andy. 2015.
\newblock \emph{Surfing uncertainty: Prediction, action, and the embodied
  mind}.
\newblock Oxford University Press.

\bibitem[{Cole(2023)}]{cole2023my}
Cole, Samantha. 2023.
\newblock ‘{M}y {AI} is sexually harassing me’: Replika users say the
  chatbot has gotten way too horny.
\newblock \emph{Motherboard: Tech by Vice}.
\newblock Https://www.vice.com/en/article/my-ai-is-sexually-harassing-me-
  replika-chatbot-nudes/ [Online. Accessed March 8,2025].

\bibitem[{Cookson et~al.(2023)Cookson, Fuentes, Kuss, and
  Bitterly}]{cookson2023social}
Cookson, T.~P., L.~Fuentes, M.~K. Kuss, and J.~Bitterly. 2023.
\newblock Social norms, gender and development: A review of research and
  practice.
\newblock Technical Report~42, {UN}-Women, New York.

\bibitem[{Cover(1999)}]{cover1999elements}
Cover, Thomas~M. 1999.
\newblock \emph{Elements of information theory}.
\newblock John Wiley \& Sons.

\bibitem[{Croskerry(2002)}]{croskerry2002achieving}
Croskerry, Pat. 2002.
\newblock Achieving quality in clinical decision making: cognitive strategies
  and detection of bias.
\newblock \emph{Academic emergency medicine}, 9(11):1184--1204.

\bibitem[{Dhodapkar(2019)}]{dhodapkar2019autoimmune}
Dhodapkar, Kavita~M. 2019.
\newblock Autoimmune complications of cancer immunotherapy.
\newblock \emph{Current opinion in immunology}, 61:54--59.

\bibitem[{Dunning(1994)}]{dunning1994accurate}
Dunning, Ted. 1994.
\newblock Accurate methods for the statistics of surprise and coincidence.
\newblock \emph{Computational linguistics}, 19(1):61--74.

\bibitem[{Ethayarajh and Jurafsky(2020)}]{ethayarajh-jurafsky-2020-utility}
Ethayarajh, Kawin and Dan Jurafsky. 2020.
\newblock Utility is in the eye of the user: A critique of {NLP} leaderboards.
\newblock In \emph{Proceedings of the 2020 Conference on Empirical Methods in
  Natural Language Processing (EMNLP)}, pages 4846--4853, Association for
  Computational Linguistics, Online.

\bibitem[{Ettinger(2020)}]{ettinger-2020-bert}
Ettinger, Allyson. 2020.
\newblock What {BERT} is not: Lessons from a new suite of psycholinguistic
  diagnostics for language models.
\newblock \emph{Transactions of the Association for Computational Linguistics},
  8:34--48.

\bibitem[{Firth(1957)}]{firth1957}
Firth, J.~R. 1957.
\newblock A synopsis of linguistic theory 1930--1955.
\newblock In \emph{Selected Papers of J. R. Firth}. Longman, London.
\newblock F. Palmer, editor.

\bibitem[{Fowler(2023)}]{fowler2023ai}
Fowler, Geoffrey~A. 2023.
\newblock {AI} is acting'pro-anorexia'and tech companies aren't stopping it.
\newblock \emph{The Washington Post}.

\bibitem[{Fridman et~al.(2019)Fridman, Barrett, Wormwood, and
  Quigley}]{fridman2019applying}
Fridman, Joseph, Lisa~Feldman Barrett, Jolie~B Wormwood, and Karen~S Quigley.
  2019.
\newblock Applying the theory of constructed emotion to police decision making.
\newblock \emph{Frontiers in Psychology}, 10:463151.

\bibitem[{Gallegos et~al.(2024)Gallegos, Rossi, Barrow, Tanjim, Kim,
  Dernoncourt, Yu, Zhang, and Ahmed}]{gallegos2024bias}
Gallegos, Isabel~O., Ryan~A. Rossi, Joe Barrow, Md~Mehrab Tanjim, Sungchul Kim,
  Franck Dernoncourt, Tong Yu, Ruiyi Zhang, and Nesreen~K. Ahmed. 2024.
\newblock Bias and fairness in large language models: A survey.
\newblock \emph{Computing Research Repository}, arXiv:2403.00770.

\bibitem[{Gallie(1955)}]{gallie1955essentially}
Gallie, Walter~Bryce. 1955.
\newblock Essentially contested concepts.
\newblock In \emph{Proceedings of the Aristotelian society}, volume~56, pages
  167--198, JSTOR.

\bibitem[{Gelman et~al.(1995)Gelman, Carlin, Stern, and
  Rubin}]{gelman1995bayesian}
Gelman, Andrew, John~B Carlin, Hal~S Stern, and Donald~B Rubin. 1995.
\newblock \emph{{Bayesian} data analysis}.
\newblock Chapman and Hall/CRC.

\bibitem[{Gigerenzer and Brighton(2009)}]{gigerenzer2009homo}
Gigerenzer, Gerd and Henry Brighton. 2009.
\newblock Homo heuristicus: Why biased minds make better inferences.
\newblock \emph{Topics in cognitive science}, 1(1):107--143.

\bibitem[{Gold(1967)}]{gold1967language}
Gold, E~Mark. 1967.
\newblock Language identification in the limit.
\newblock \emph{Information and control}, 10(5):447--474.

\bibitem[{Gopnik and Wellman(2012)}]{gopnik2012reconstructing}
Gopnik, Alison and Henry~M Wellman. 2012.
\newblock Reconstructing constructivism: causal models, {B}ayesian learning
  mechanisms, and the theory theory.
\newblock \emph{Psychological bulletin}, 138(6):1085.

\bibitem[{Gordon(2017)}]{Gordon2017}
Gordon, Maggie. 2017.
\newblock "me too" the "end of the beginning" of a movement: Many now wrestling
  with how to turn a hashtag into real-life change.
\newblock \emph{Houston Chronicle}.
\newblock Accessed: 2025-03-06.

\bibitem[{Graber-Stiehl(2023)}]{graber2023world}
Graber-Stiehl, Ian. 2023.
\newblock Is the world ready for {ChatGPT} therapists?
\newblock \emph{Nature}, 617(7959):22--24.

\bibitem[{Groeneveld et~al.(2024)Groeneveld, Beltagy, Walsh, Bhagia, Kinney,
  Tafjord, Jha, Ivison, Magnusson, Wang, Arora, Atkinson, Authur, Chandu,
  Cohan, Dumas, Elazar, Gu, Hessel, Khot, Merrill, Morrison, Muennighoff, Naik,
  Nam, Peters, Pyatkin, Ravichander, Schwenk, Shah, Smith, Strubell, Subramani,
  Wortsman, Dasigi, Lambert, Richardson, Zettlemoyer, Dodge, Lo, Soldaini,
  Smith, and Hajishirzi}]{groeneveld2024olmo}
Groeneveld, Dirk, Iz~Beltagy, Pete Walsh, Akshita Bhagia, Rodney Kinney, Oyvind
  Tafjord, Ananya~Harsh Jha, Hamish Ivison, Ian Magnusson, Yizhong Wang, Shane
  Arora, David Atkinson, Russell Authur, Khyathi~Raghavi Chandu, Arman Cohan,
  Jennifer Dumas, Yanai Elazar, Yuling Gu, Jack Hessel, Tushar Khot, William
  Merrill, Jacob Morrison, Niklas Muennighoff, Aakanksha Naik, Crystal Nam,
  Matthew~E. Peters, Valentina Pyatkin, Abhilasha Ravichander, Dustin Schwenk,
  Saurabh Shah, Will Smith, Emma Strubell, Nishant Subramani, Mitchell
  Wortsman, Pradeep Dasigi, Nathan Lambert, Kyle Richardson, Luke Zettlemoyer,
  Jesse Dodge, Kyle Lo, Luca Soldaini, Noah~A. Smith, and Hannaneh Hajishirzi.
  2024.
\newblock {OLMo}: Accelerating the science of language models.
\newblock \emph{Computing Research Repository}, arXiv:2402.00838.
\newblock Version v3.

\bibitem[{Grossman(2024)}]{grossman2024tech}
Grossman, Gary. 2024.
\newblock Tech’s new arms race: The billion-dollar battle to build {AI}.
\newblock \emph{Venture{B}eat}.
\newblock [Online. Accessed May 15, 2024].

\bibitem[{Halevy, Norvig, and Pereira(2009)}]{halevy2009unreasonable}
Halevy, Alon, Peter Norvig, and Fernando Pereira. 2009.
\newblock The unreasonable effectiveness of data.
\newblock \emph{{IEEE} intelligent systems}, 24(2):8--12.

\bibitem[{Harnad(1990)}]{harnad1990symbol}
Harnad, Stevan. 1990.
\newblock The symbol grounding problem.
\newblock \emph{Physica D: Nonlinear Phenomena}, 42(1-3):335--346.

\bibitem[{Harris(1954)}]{harris1954distributional}
Harris, Zellig~S. 1954.
\newblock Distributional structure.
\newblock \emph{Word}, 10(2-3):146--162.

\bibitem[{Hastie et~al.(2009)Hastie, Tibshirani, Friedman, and
  Friedman}]{hastie2009elements}
Hastie, Trevor, Robert Tibshirani, Jerome~H Friedman, and Jerome~H Friedman.
  2009.
\newblock \emph{The Elements of Statistical Learning: Data Mining, Inference,
  and Prediction}, volume~2.
\newblock Springer.

\bibitem[{Haussler(1988)}]{haussler1988quantifying}
Haussler, David. 1988.
\newblock Quantifying inductive bias: {AI} learning algorithms and valiant's
  learning framework.
\newblock \emph{Artificial intelligence}, 36(2):177--221.

\bibitem[{Hirota, Nakashima, and Garcia(2022)}]{hirota2022quantifying}
Hirota, Yusuke, Yuta Nakashima, and Noa Garcia. 2022.
\newblock Quantifying societal bias amplification in image captioning.
\newblock In \emph{Proceedings of the {IEEE/CVF} conference on computer vision
  and pattern recognition}, pages 13450--13459.

\bibitem[{Hofmann et~al.(2024)Hofmann, Kalluri, Jurafsky, and
  King}]{hofmann2024dialect}
Hofmann, Valentin, Pratyusha~Ria Kalluri, Dan Jurafsky, and Sharese King. 2024.
\newblock Dialect prejudice predicts {AI} decisions about people's character,
  employability, and criminality.
\newblock \emph{Computing Research Repository}, arXiv:2403.00742.

\bibitem[{Iskander(2022)}]{iskander2022manhattan}
Iskander, George. 2022.
\newblock The {M}anhattan project shows scientists’ moral and ethical
  responsibilities.
\newblock \emph{Scientific American}.

\bibitem[{Jelinek, Bahl, and Mercer(1975)}]{jelinek1975design}
Jelinek, Frederick, Lalit Bahl, and Robert Mercer. 1975.
\newblock Design of a linguistic statistical decoder for the recognition of
  continuous speech.
\newblock \emph{{IEEE} Transactions on Information Theory}, 21(3):250--256.

\bibitem[{Jurafsky and Martin(2024)}]{jurafsky3rd}
Jurafsky, Dan and James~H Martin. 2024.
\newblock Speech and language processing.
\newblock 3rd edition draft, February 3, 2024.

\bibitem[{Kaeser-Chen et~al.(2020)Kaeser-Chen, Dubois, Sch{\"u}{\"u}r, and
  Moss}]{kaeser2020positionality}
Kaeser-Chen, Christine, Elizabeth Dubois, Friederike Sch{\"u}{\"u}r, and
  Emanuel Moss. 2020.
\newblock Positionality-aware machine learning: translation tutorial.
\newblock In \emph{Proceedings of the 2020 Conference on fairness,
  accountability, and transparency}, pages 704--704.

\bibitem[{Knill and Pouget(2004)}]{knill2004bayesian}
Knill, David~C and Alexandre Pouget. 2004.
\newblock The {B}ayesian brain: the role of uncertainty in neural coding and
  computation.
\newblock \emph{TRENDS in Neurosciences}, 27(12):712--719.

\bibitem[{Lambert et~al.(2022)Lambert, Castricato, von Werra, and
  Havrilla}]{lambert2022illustrating}
Lambert, N, L~Castricato, L~von Werra, and A~Havrilla. 2022.
\newblock Illustrating reinforcement learning from human feedback ({RLHF}).
\newblock [Online; accessed April 27, 2024].

\bibitem[{Landauer and Dumais(1997)}]{landauer1997solution}
Landauer, Thomas~K and Susan~T Dumais. 1997.
\newblock A solution to {P}lato's problem: The latent semantic analysis theory
  of acquisition, induction, and representation of knowledge.
\newblock \emph{Psychological review}, 104(2):211.

\bibitem[{Lee and Seung(2000)}]{lee2000algorithms}
Lee, Daniel and H~Sebastian Seung. 2000.
\newblock Algorithms for non-negative matrix factorization.
\newblock \emph{Advances in neural information processing systems}, 13.

\bibitem[{Lenci(2018)}]{lenci2018distributional}
Lenci, Alessandro. 2018.
\newblock Distributional models of word meaning.
\newblock \emph{Annual Review of Linguistics}, 4:151--171.

\bibitem[{Linzen and Baroni(2021)}]{linzen2021syntactic}
Linzen, Tal and Marco Baroni. 2021.
\newblock Syntactic structure from deep learning.
\newblock \emph{Annual Review of Linguistics}, 7:195--212.

\bibitem[{Mahowald et~al.(2024)Mahowald, Ivanova, Blank, Kanwisher, Tenenbaum,
  and Fedorenko}]{mahowald2024dissociating}
Mahowald, Kyle, Anna~A Ivanova, Idan~A Blank, Nancy Kanwisher, Joshua~B
  Tenenbaum, and Evelina Fedorenko. 2024.
\newblock Dissociating language and thought in large language models.
\newblock \emph{Trends in Cognitive Sciences}, 28:517--540.

\bibitem[{Manning and Sch\"utze(1999)}]{manning1999foundations}
Manning, Christopher and Hinrich Sch\"utze. 1999.
\newblock \emph{Foundations of Statistical Natural Language Processing}.
\newblock MIT Press.

\bibitem[{Markov(1913)}]{markov1913}
Markov, A.~A. 1913.
\newblock Essai d'une recherche statistique sur le texte du roman ``{E}ugene
  {O}negin'' illustrant la liaison des epreuve en chain ({`Example} of a
  statistical investigation of the text of ``{E}ugene {O}negin" illustrating
  the dependence between samples in chain').
\newblock \emph{Izvistia Imperatorskoi Akademii Nauk (Bulletin de
  {l'Acad\'{e}mie} {Imp\'{e}riale} des Sciences de {St.-P\'{e}tersbourg)}},
  7:153--162.
\newblock English translation by Morris Halle, 1956.

\bibitem[{Maslej et~al.(2024)Maslej, Fattorini, Perrault, Parli, Reuel,
  Brynjolfsson, Etchemendy, Ligett, Lyons, Manyika, Niebles, Shoham, Wald, and
  Clark}]{Maslej2024AIIndex}
Maslej, Nestor, Loredana Fattorini, Raymond Perrault, Vanessa Parli, Anka
  Reuel, Erik Brynjolfsson, John Etchemendy, Katrina Ligett, Terah Lyons, James
  Manyika, Juan~Carlos Niebles, Yoav Shoham, Russell Wald, and Jack Clark.
  2024.
\newblock The {AI} index 2024 annual report.
\newblock {AI} index report, {AI} Index Steering Committee, Institute for
  Human-Centered {AI}, Stanford University, Stanford, CA.
\newblock [Online; accessed May 4, 2024].

\bibitem[{Metz and Love(2024)}]{metz2024}
Metz, Rachel and Julia Love. 2024.
\newblock Character.{AI} co-founders hired by google in licensing deal.
\newblock \emph{Bloomberg}.
\newblock Published via Yahoo Finance.

\bibitem[{Michael(2020)}]{julianmichael}
Michael, Julian. 2020.
\newblock To dissect an octopus: Making sense of the form/meaning debate.
\newblock [Online; Accessed March 27, 2024].

\bibitem[{Michaelov et~al.(2024)Michaelov, Bardolph, Petten, Bergen, and
  Coulson}]{Michaelov2023StrongPL}
Michaelov, James~A., Megan~D. Bardolph, Cyma K.~Van Petten, Benjamin~K. Bergen,
  and Seana Coulson. 2024.
\newblock Strong prediction: Language model surprisal explains multiple {N400}
  effects.
\newblock \emph{Neurobiology of Language}, pages 1--29.
\newblock Advance publication.

\bibitem[{Mitchell(1980)}]{mitchell1980need}
Mitchell, Tom~M. 1980.
\newblock The need for biases in learning generalizations.
\newblock Technical Report CBM-TR-117, Department of Computer Science,
  Laboratory for Computer Science Research, Rutgers University.

\bibitem[{Nagel(1989)}]{nagel1989view}
Nagel, Thomas. 1989.
\newblock \emph{The view from nowhere}.
\newblock Oxford University Press.

\bibitem[{{National Conference of State Legislatures}(2019)}]{NCSL2019}
{National Conference of State Legislatures}. 2019.
\newblock Legislation on sexual harassment in the legislature.
\newblock Version of February 11, 2019. [Online. Accessed March 6, 2025].

\bibitem[{Ng et~al.(2025)Ng, Goh, Teo, Chong, Tan, and Teoh}]{ng2025clinical}
Ng, Isaac~KS, Wilson~GW Goh, Desmond~B Teo, Kar~Mun Chong, Li~Feng Tan, and
  Chia~Meng Teoh. 2025.
\newblock Clinical reasoning in real-world practice: a primer for medical
  trainees and practitioners.
\newblock \emph{Postgraduate Medical Journal}, 101(1191):68--75.

\bibitem[{Nitoburg(1933)}]{nitoberg1933}
Nitoburg, Lev. 1933.
\newblock \emph{The German Quarter}.
\newblock {R}ussia: {S}oviet {L}iterature ({S}ovetskya {L}iteratura).

\bibitem[{Ouyang et~al.(2022)Ouyang, Wu, Jiang, Almeida, Wainwright, Mishkin,
  Zhang, Agarwal, Slama, Ray et~al.}]{ouyang2022training}
Ouyang, Long, Jeffrey Wu, Xu~Jiang, Diogo Almeida, Carroll Wainwright, Pamela
  Mishkin, Chong Zhang, Sandhini Agarwal, Katarina Slama, Alex Ray, et~al.
  2022.
\newblock Training language models to follow instructions with human feedback.
\newblock \emph{Advances in neural information processing systems},
  35:27730--27744.

\bibitem[{Rattner and Dean(2025)}]{rattner2025ai}
Rattner, Nate and Jason Dean. 2025.
\newblock Tech giants double down on their massive {AI} spending.
\newblock \emph{Wall Street Journal}.
\newblock Accessed online March 8, 2025.

\bibitem[{Reason(1990)}]{reason1990human}
Reason, James. 1990.
\newblock \emph{Human error}.
\newblock Cambridge University Press.

\bibitem[{Reiss and Sprenger(2020)}]{sep-scientific-objectivity}
Reiss, Julian and Jan Sprenger. 2020.
\newblock {Scientific Objectivity}.
\newblock In Edward~N. Zalta, editor, \emph{The {Stanford} Encyclopedia of
  Philosophy}, {W}inter 2020 edition. Metaphysics Research Lab, Stanford
  University.

\bibitem[{Rodriguez(2020)}]{Rodriguez2020}
Rodriguez, Adrianna. 2020.
\newblock Goodbye, handshake. {H}ello, elbow bump? {G}reetings to avoid during
  the coronavirus outbreak.
\newblock \emph{USA Today}.
\newblock Retrieved 2025-03-06.

\bibitem[{Roose(2024)}]{roose2024}
Roose, Kevin. 2024.
\newblock Can {A.I.} be blamed for a teen’s suicide?
\newblock \emph{The New York Times}.

\bibitem[{Roose and Newton(2024)}]{hardfork2024}
Roose, Kevin and Casey Newton. 2024.
\newblock The {E}lon-ction + can {A.I.} be blamed for a teen's suicide?
\newblock Podcast episode.

\bibitem[{Roth(2024)}]{roth2024google}
Roth, Emma. 2024.
\newblock Google explains {G}emini’s ‘embarrassing’ {AI} pictures of
  diverse {N}azis.
\newblock \emph{The Verge}.
\newblock [Online. Accessed May 15, 2024].

\bibitem[{Sap et~al.(2019)Sap, Card, Gabriel, Choi, and
  Smith}]{sap-etal-2019-risk}
Sap, Maarten, Dallas Card, Saadia Gabriel, Yejin Choi, and Noah~A. Smith. 2019.
\newblock The risk of racial bias in hate speech detection.
\newblock In \emph{Proceedings of the 57th Annual Meeting of the Association
  for Computational Linguistics}, pages 1668--1678, Association for
  Computational Linguistics, Florence, Italy.

\bibitem[{Saphra et~al.(2024)Saphra, Fleisig, Cho, and
  Lopez}]{saphra2024tragedy}
Saphra, Naomi, Eve Fleisig, Kyunghyun Cho, and Adam Lopez. 2024.
\newblock First tragedy, then parse: History repeats itself in the new era of
  large language models.
\newblock \emph{Computing Research Repository}, arXiv:2311.05020.

\bibitem[{Schwartz et~al.(2021)Schwartz, Down, Jonas, and
  Tabassi}]{schwartz2021proposal}
Schwartz, Reva, Leann Down, Adam Jonas, and Elham Tabassi. 2021.
\newblock A proposal for identifying and managing bias in artificial
  intelligence.
\newblock \emph{Draft NIST Special Publication}, 1270.

\bibitem[{Schwartz(2023)}]{schwartz2023}
Schwartz, Steven~A. 2023.
\newblock Sworn statement in {Roberto Mata} v {Avianca Inc}.
\newblock United States District Court for the Southern District of New York,
  Civil Action No. 22-cv-1461.

\bibitem[{Serrano, Dodge, and Smith(2023)}]{serrano-etal-2023-stubborn}
Serrano, Sofia, Jesse Dodge, and Noah~A. Smith. 2023.
\newblock Stubborn lexical bias in data and models.
\newblock In \emph{Findings of the Association for Computational Linguistics:
  ACL 2023}, pages 8131--8146, Association for Computational Linguistics,
  Toronto, Canada.

\bibitem[{Shannon(1948)}]{shannon1948mathematical}
Shannon, Claude~Elwood. 1948.
\newblock A mathematical theory of communication.
\newblock \emph{The Bell system technical journal}, 27(3):379--423.

\bibitem[{Simon(1956)}]{simon1956rational}
Simon, Herbert~A. 1956.
\newblock Rational choice and the structure of the environment.
\newblock \emph{Psychological review}, 63(2):129.

\bibitem[{Simon(1969)}]{simon1969}
Simon, Herbert~A. 1969.
\newblock \emph{The sciences of the artificial}.
\newblock Massachusetts Institute of Technology.

\bibitem[{Tenenbaum et~al.(2011)Tenenbaum, Kemp, Griffiths, and
  Goodman}]{tenenbaum2011grow}
Tenenbaum, Joshua~B, Charles Kemp, Thomas~L Griffiths, and Noah~D Goodman.
  2011.
\newblock How to grow a mind: Statistics, structure, and abstraction.
\newblock \emph{science}, 331(6022):1279--1285.

\bibitem[{Vance(2025)}]{vance2025ai}
Vance, J.~D. 2025.
\newblock Remarks by the {V}ice {P}resident at the artificial intelligence
  action summit in {P}aris, {F}rance [transcript].
\newblock The American Presidency Project.

\bibitem[{VandenBos(2007)}]{vandenbos2007apa}
VandenBos, Gary~R. 2007.
\newblock \emph{{APA} dictionary of psychology.}
\newblock American Psychological Association.
\newblock Online. Accessed October 9, 2024.

\bibitem[{Wikipedia(2024)}]{wikipedia:frog}
Wikipedia. 2024.
\newblock The {S}corpion and the {Frog}.
\newblock [Online; accessed 6 April 2024. Cites Nitoberg (1933)].

\bibitem[{Wilcox et~al.(2020)Wilcox, Gauthier, Hu, Qian, and
  Levy}]{wilcox2020predictive}
Wilcox, Ethan~Gotlieb, Jon Gauthier, Jennifer Hu, Peng Qian, and Roger Levy.
  2020.
\newblock On the predictive power of neural language models for human real-time
  comprehension behavior.
\newblock Proceedings of the Annual Meeting of the Cognitive Science Society.

\bibitem[{Williams et~al.(2024)Williams, Burke-Moore, Chan, Enock, Nanni,
  Sippy, Chung, Gabasova, Hackenburg, and Bright}]{williams2024large}
Williams, Angus~R, Liam Burke-Moore, Ryan Sze-Yin Chan, Florence~E Enock,
  Federico Nanni, Tvesha Sippy, Yi-Ling Chung, Evelina Gabasova, Kobi
  Hackenburg, and Jonathan Bright. 2024.
\newblock Large language models can consistently generate high-quality content
  for election disinformation operations.
\newblock \emph{arXiv preprint arXiv:2408.06731}.

\bibitem[{Writer(2023)}]{neda2023}
Writer, Staff. 2023.
\newblock {NEDA} suspends {AI} chatbot for giving harmful eating disorder
  advice.
\newblock \emph{{Psychiatrist.com}}.

\end{thebibliography}

\end{document}

}